\documentclass{article} 
\usepackage{epsfig}

\newcommand{\Fmax}{ {F_{\rm max}} }
\newcommand{\Fmin}{ {F_{\rm min}} }
\newcommand{\pmax}{ {p_{\rm max}} }
\newcommand{\pmin}{ {p_{\rm min}} }

\begin{document}

\title{Multiagent Control of Self-reconfigurable Robots\thanks{A condensed version of this paper will appear in Proc. of the Intl. Conf. on Multiagent Systems (ICMAS2000), July 2000.}}

\author{Hristo Bojinov, Arancha Casal and Tad Hogg        \\
        Xerox Palo Alto Research Center\\
	3333 Coyote Hill Road\\
	Palo Alto, CA 94304}
\maketitle

\begin{abstract}
We demonstrate how multiagent systems provide useful control
techniques for modular self-reconfigurable (metamorphic) robots. Such
robots consist of many modules that can move relative to each other,
thereby changing the overall shape of the robot to suit different
tasks. Multiagent control is particularly well-suited for tasks
involving uncertain and changing environments. We illustrate this
approach through simulation experiments of Proteo, a metamorphic robot
system currently under development.
\end{abstract}

\section{Introduction}

Modular self-reconfigurable (or metamorphic)
robots~\cite{yim94,pamecha97,yim97,rus99,kotay98a,murata94,murata98}
consist of many simple identical modules, that can attach and detach
from one another to change their overall topology. These systems can
dynamically adapt their shape to suit the needs of the task at hand,
e.g., for manipulation, locomotion, and adaptive structures responding
to environmental stresses.

From a planning and control viewpoint, metamorphic robots pose several
interesting research challenges. Self-reconfiguration, or how to
change shape automatically, is a new and so far little studied problem
for robots. Decentralized control is a useful approach to this problem
especially when the robot has a the large number of modules, each of
which is a self-contained unit with its own processing, sensing and
actuation.  With the tight physical interactions due to contact
between neighboring modules and constraints arising from actuator
geometry and power limitations, modular metamorphic robots pose an
interesting challenge for multiagent control. They thus require a more
physically grounded approach than many studies of artificial
life~\cite{epstein96} which tend to concentrate on how complex natural
organisms achieve sophisticated crowd behaviors or deal with abstract
agents that currently can not be physically
constructed~\cite{hackwood92,hall96,abelson99}.  Alternatively,
biological and chemical techniques for distributed construction of
shapes~\cite{bowden97,metzger99} provide examples of local behaviors
producing complex shapes, but are not yet capable of producing general
programmable robotic systems.

One approach to reconfiguration uses a precise specification of the
desired locations of all the modules, and then solves the
combinatorial search required to identify motions for the modules
according to some criterion, such as minimizing the number of moves or
power consumption. Such searches are generally intractable for robots
consisting of many modules, but can be addressed approximately using
heuristics~\cite{pamecha97,yim97,murata98,kotay98a,rus99,casal99,hosokawa98}.
Unfortunately, in many practical applications, defining an exact
target shape may not be suitable or even possible. This may arise when
some modules fail or the nature of the environment or task is
uncertain, for example, when grasping an object of unknown size or
shape.

Instead, we use multiagent control to achieve suitable reconfiguration
as a side-effect of creating a structure with the {\em properties}
(structural, morphological, etc) required for the task. When the
properties can be expressed largely in terms of the local environment
for each module, agent-based control often achieves a suitable shape
without any need to precisely specify the exact position of each
module.  Moreover, an agent-based architecture is well-suited to
decomposing control problems based on the different physical phenomena
dominant at different scales, especially for metamorphic robots with a
large number of tiny modules. For example, micromachined
robots~\cite{ebefors99} are dominated by friction and other surface
forces rather than gravity.  Even smaller structures~\cite{howard97}
are subject to randomly fluctuating forces, i.e., Brownian motion. In
contrast to larger robots, these tiny machines readily move objects
many times their own weight and have comparatively high speeds and
strengths~\cite{drexler92}.  Biology has numerous examples of
different structures appropriate for different
scales~\cite{thompson92}.  In such cases, different types of agents
could be responsible for behaviors of individual modules, small groups
of modules and so on, forming a hierarchical or multihierarchical
correspondence between physical structures of the robot and its
environment, the levels of specification for the desired task and the
controlling software~\cite{hogg96b}.

A related line of work applies multiagent control to teams of robots
cooperating to achieve a common
task~\cite{steels89,caloud90,Rush94,hasslacher95,kitano98}. These
methods usually apply to independently mobile robots that are not
physically connected and have little or no physical contact. In most
current modular robot systems the modules remain attached to one
another forming a single connected
whole~\cite{casal99,rus99,pamecha97,murata94,murata98,yim97}, and
giving rise to a number of tight physical motion constraints that do
not apply to teams of independent robots.  On the other hand, the
physical contact between modules allows modules to locate their
neighbors without complex sensory processing as would be required, for
example, to visually identify a physically disconnected member of a
team. Hence the techniques for coordinating teams are not directly
applicable to modular robot controls.  Furthermore, more traditional
self-reconfiguration algorithms require an a-priori exact description
of a target shape for the given task, which may not be suitable when
the robot operates in uncertain environments.

Specifically, we explore the use of {\em simple}, purely {\em local}
rules to produce control algorithms for accomplishing tasks such as
dynamic adaptation under changing external conditions (e.g., added
weight) and grasping objects. We will use the term {\em behaviors} for
the specific control algorithms. The assumptions we make are as
follows:

\begin{itemize}

\item{Modules have limited computational capabilities, consisting of a
limited memory and a simple finite-state machine (FSM). State
transitions are driven by the states, the relative positions of a
module and its neighbors, and some external sensor information. The
state transitions can update the memory of a module and its
neighbors.}

\item{Communication is limited to immediate neighbors, and a limited number
of bits are exchanged at each step. We specifically did not allow
modules to broadcast messages globally, because the power required
would likely not scale well if the size of modules shrinks. Morover,
fully utilizing broadcasts would require more complex knowledge and
processing on the part of the modules.}
\end{itemize}

The remainder of this paper presents the design and evaluation of
control algorithms for metamorphic systems that coordinate their
actions locally to achieve emergent, global behaviors. The next
section describes the robot platform used in our work and general
control primitives. The following two sections present the results of
our algorithms applied to specific tasks and provide an analysis of
the behavior.

\section{The Robot System}

This section describes the modular robot used in our work, its
simulator and the control primitives combined to create the behaviors
described below.

\subsection{Experimental Robot Platform}

Several modular metamorphic robot designs have been
proposed~\cite{fukuda88,rus99,kotay98a,pamecha97,murata94,murata98,yim94}.
For our study, we focus on one such design: Proteo~\cite{yim97}, which
is a modular self-reconfigurable, or metamorphic, robot being
developed at Xerox PARC. In the current design, modules are rhombic
dodecahedra (polygons with twelve identical faces, each of which is a
rhombus). Modules attach to one another along their faces forming
general three dimensional solids.  To achieve a change of shape
modules on the outer surface roll over the substrate of other modules
to new positions. This type of reconfiguration has been called
``substrate reconfiguration'', in contrast with other existing
reconfiguration classes~\cite{casal99}.

\begin{figure}[ht]
\begin{center}
\epsfig{file=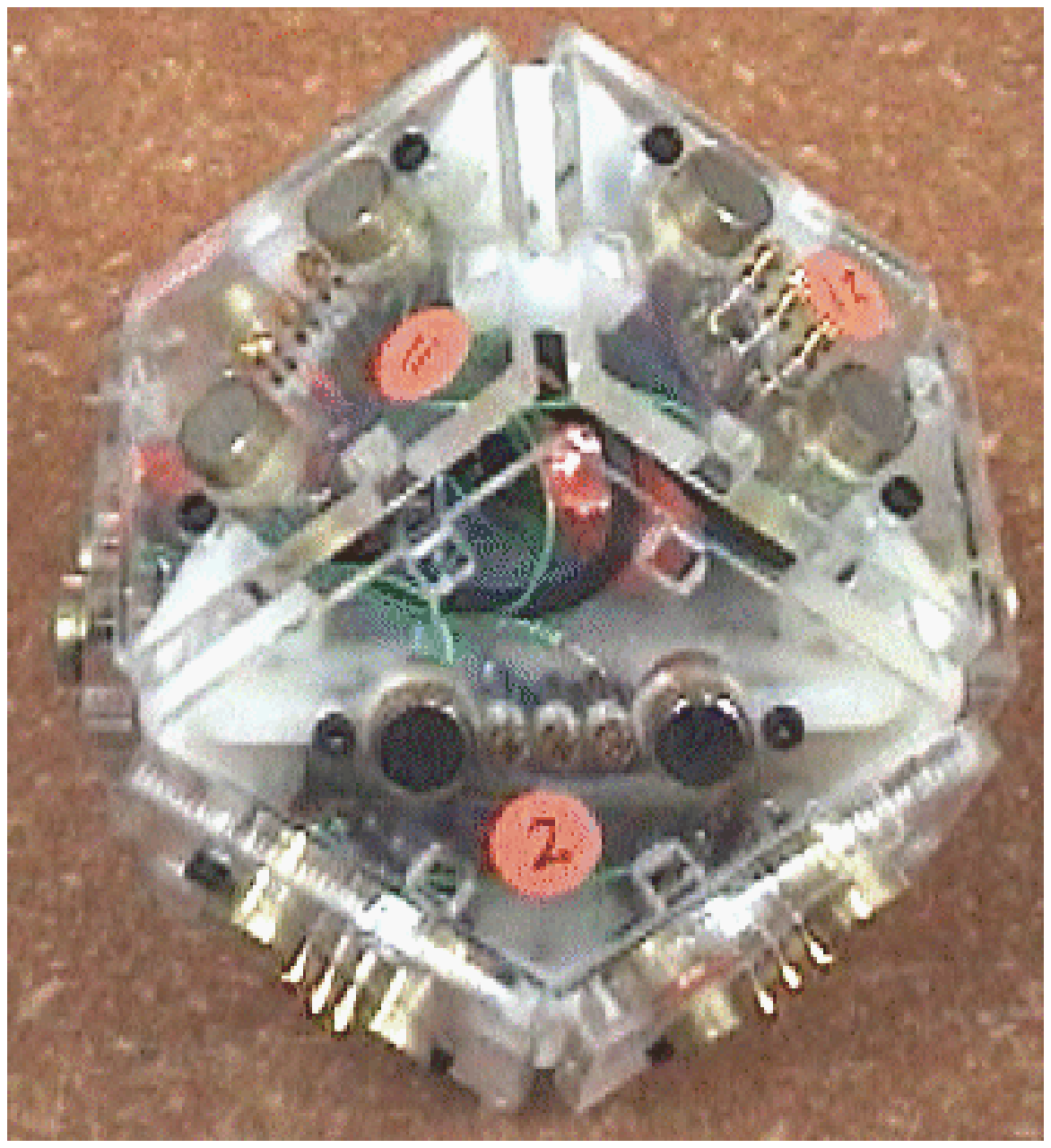,width=166pt}  
\epsfig{file=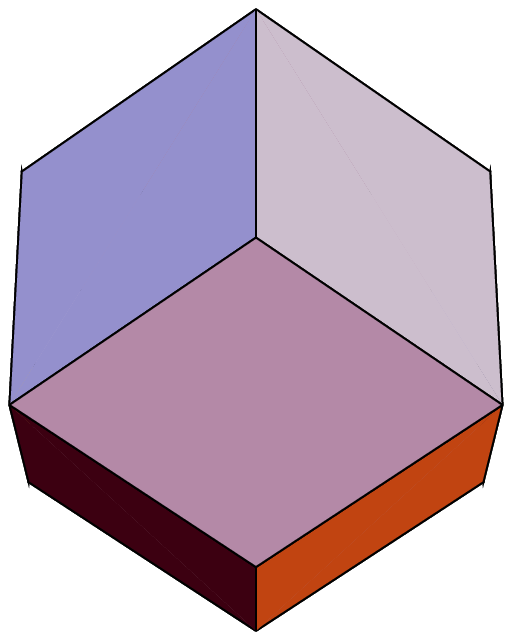,width=166pt}
\end{center}
\caption{An actual proteo module and its schematic representation.}\label{proteo}
\end{figure}

\begin{figure}[htb]
\begin{center}
\epsfig{file=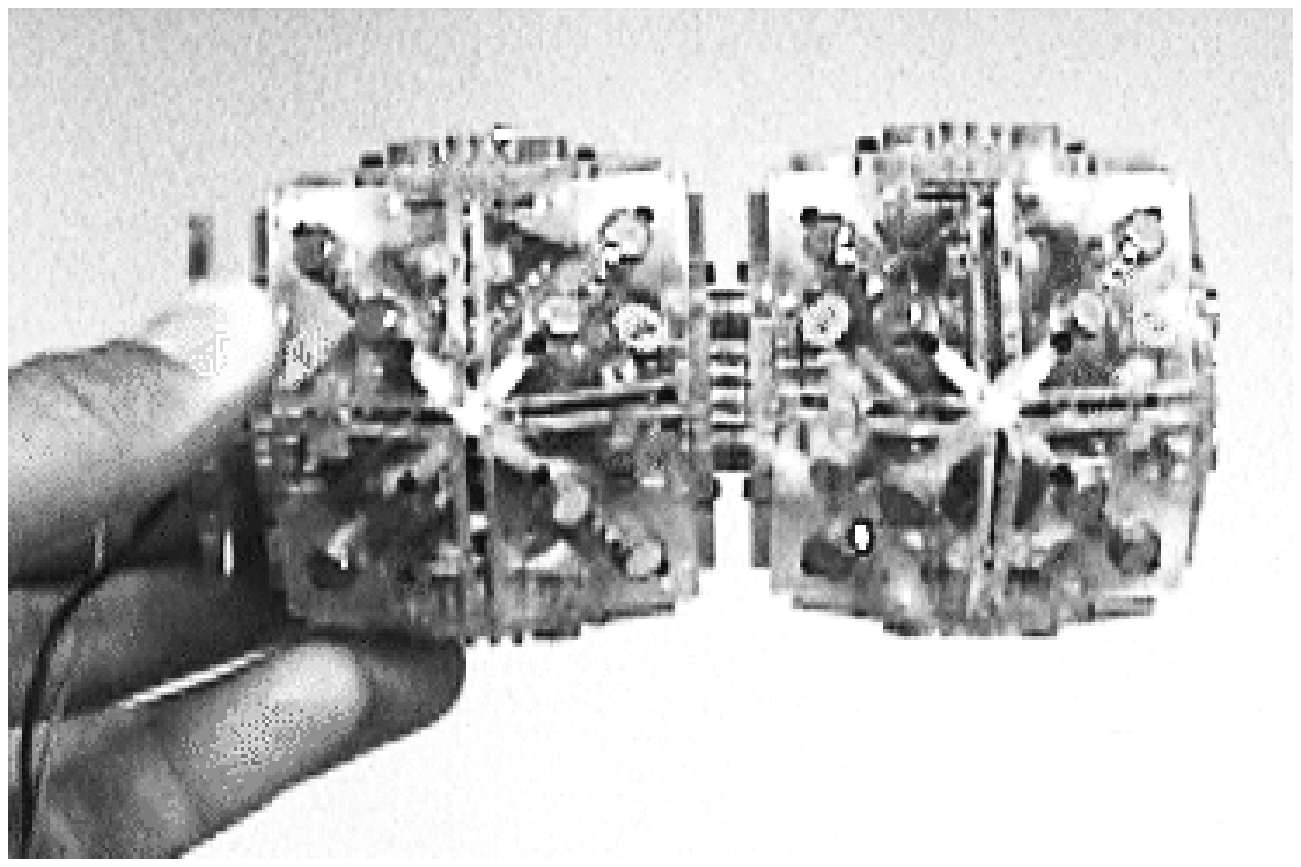,width=185pt}  
\epsfig{file=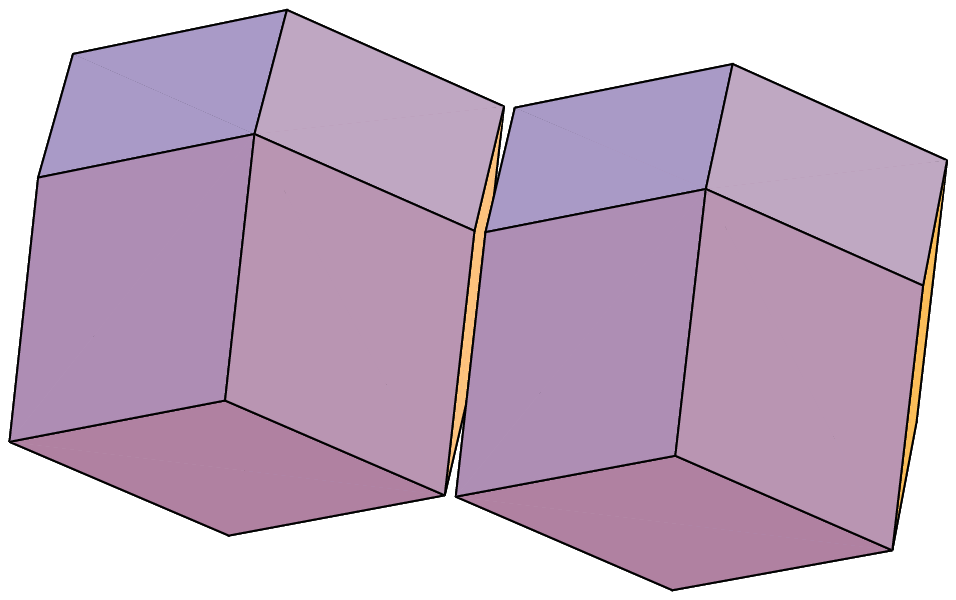,width=185pt}
\end{center}
\caption{Two connected proteo modules and their schematic representation from a different viewpoint. Each module can rotate onto its neighbor's adjacent faces.}\label{proteo2}
\end{figure}

The rhombic dodecahedron can be thought of as the 3D analog of the
hexagon and has several useful properties. Notably, these polyhedra
result in maximium internal volume for a given surface
area~\cite{thompson92}, meaning more room for packing electronic and
mechanical components. Also, all module-on-module rotations are always
120 degrees, unlike the cube which requires 180 degree rotations in
certain cases. A main disadvantage is that twelve faces per module
need to be actuated, increasing the complexity and expense of the
hardware. For a target module size in the centimeter scale, current
actuator technologies result in large weight to power ratios and high
cost per module. A manually actuated version of Proteo has been built
and is shown in Figure~\ref{proteo}.  Modules connect along their
faces, as illustrated in Figure~\ref{proteo2}.

In addition to actuators, the modules can communicate directly with
their neighbors. Depending on the application, they can also include
various sensor elements, e.g., to detect forces imposed by the
environment or the weight of other modules.

For the experiments reported below, we used a simulation of the Proteo
robot platform. This simulation includes the physical motion
constraints on the modules, allowing them to rotate onto neighboring
modules only when that motion is not obstructed by other modules.
This simulation has the modules operating
asynchronously. Specifically, every module is given a chance to
execute its behavior exactly once each time cycle. The behavior
executions for the modules are ordered in a different, random way for
each time cycle. When the behavior code is executed each module
describes what movement (if any) it wants to perform at the end of the
cycle. After all the behavior code is executed, the simulator attempts
to perform the requested movement for each module in turn, subject to
geometrical constraints.  This means that a module might be denied the
movement it has requested.

The control programs in the modules are not directly notified of the
outcome of their movement requests (by means of an exception, for
example). While this capability could certainly be added to the
system, we found it is not necessary for applying our coordination
techniques to the tasks described below. This lack of notification
means modules can become ``stuck'' and attempt the same movement in
vain for several time cycles. However, the overall behavior is still
achieved because at the same time other modules are able to complete
their moves, and can eventually remove the obstacle preventing
motion. Alternatively, the motion of these other modules can
sufficiently change the structure that the stuck module decides on
another motion.

Another programming issue was how a module should sample randomly from
the available moves. When a completely random move is desired, the
module simply picks with equal probability from the available moves to
neighboring positions. When a directed random move is needed, a
different method is used, which picks only from the available moves
that go in a direction that has a positive dot product with a
specified general direction in space. The latter approach is used when
there is some bias direction inherent to the specific application
(e.g. the direction of the ``ball'' in the grasping behavior described
in Section~\ref{grasp}). While such biased choices do not necessarily
result in the most direct motion toward the goal, the randomness
reduces the likelihood of becoming stuck and increases robustness in
case of failed modules. From a global planning perspective, this
randomness can be viewed as a simple search procedure that avoids the
potentially intractable combinatorial search of finding an optimal set
of motions for all the modules.

The simulation does not model all the details of the hardware. For
example, it does not include limits on power consumption or actuator
strength that could be significant in actual hardware
implementations. The simulator also assumes the modules operate
correctly and have accurate local sensory information, e.g.,
concerning contact with neighbors or other objects or mechanical
stress.  The results of the simulation are illustrated using the
schematic representation of the modules included in
Figures~\ref{proteo} and \ref{proteo2}.

\subsection{Control Primitives}

While the Proteo system illustrates the issues facing modular robotics
and provides a useful testbed, our methods are not tied to any
specific design, but instead are applicable to metamorphic robots in
general. Specifically, the agents make use of a few primitive
constructs that are combined to form a control program for each
module. These primitives allow simple communication and coordination
among the modules, and could be implemented by a variety of modular
robots.

Our approach uses the following primitives:

\begin{description}
\item{{\bf Growth}} is the process that creates structures. It is important 
to provide a mechanism for focusing the movement of modules to
specific spots. Otherwise, if the movement is too random, it could
take a long time before any reasonably good structure is grown.

\item{{\bf Seeds}} 
are the main agents that cause growth. A seed is a module that
attracts other modules in order to further grow the structure. As more
modules are attracted to a seed, they can in turn become seeds
themselves, and thus propagate the growth process.

\item{{\bf Scents}}
 are the means of global communication among modules. Scents are
 propagated through the system in a distributed breadth-first fashion
 as follows. Each module keeps track of the scent strength at its
 location with a value in its memory. Those modules that emit the
 scent set their own value to zero. Other modules examine the scent
 values of their neighbors at each time step, and then set their own
 scent value to one more than the minimum value among their neighbors.
 Thus, smaller values indicate a stronger scent, and a scent gradient
 is created throughout the structure. Scent values are an
 approximation of the minimum distance to a scent-emitting module in
 the system at any given time. The approximation arises because scents
 are broadcast via neighbors and thus can take many steps to propagate
 throughout the entire system, and during this time modules can
 move. Nevertheless, as discussed further in Section~\ref{stability},
 the scents generally propagate much faster than substantial
 rearrangements of the modules so this approximation worked quite
 well.  We typically use either one or two different scents in our
 experiments. Seeds emit a scent to attract modules, and modules that
 are searching for seeds move along the surface gradient of the scent
 to find them.

\item{The {\bf Mode}}
of a module is its present FSM state. For instance, seeds are usually
denoted by the SEED mode, modules that search for seeds are in the
SEARCH mode, modules that are part of the finished structure are in
the FINAL mode, and modules that cause branching are in NODE mode. The
mode in turn will determine the rules of behavior of a module.
\end{description}

A final component of our control method is the use of random motions
when different choices appear equally useful. This allows modules to
continue performing even when the scent gradients do not completely
specify the best direction. Because such random motions are less efficient
than following gradients, our algorithms arrange for relatively few states
to search before useful gradient information is found. 

\begin{figure}
\begin{center}
\epsfig{file=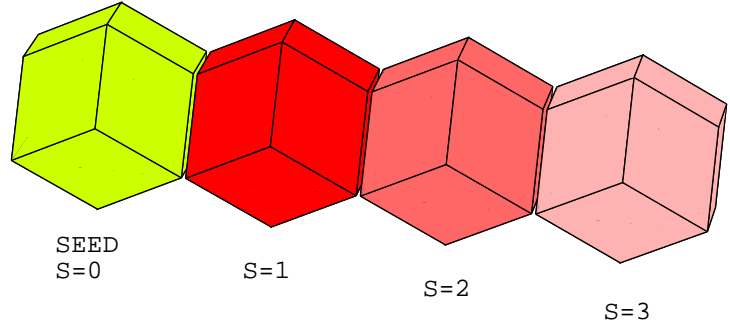,width=4in}
\end{center}
\caption{\label{search}A simple chain of modules with one SEED (yellow) and three SEARCH modules (red, with the color intensity corresponding to the scent values). In this simple structure, scent values exactly correspond to the distance from the seed.}
\end{figure}

\begin{figure}
\begin{center}
\epsfig{file=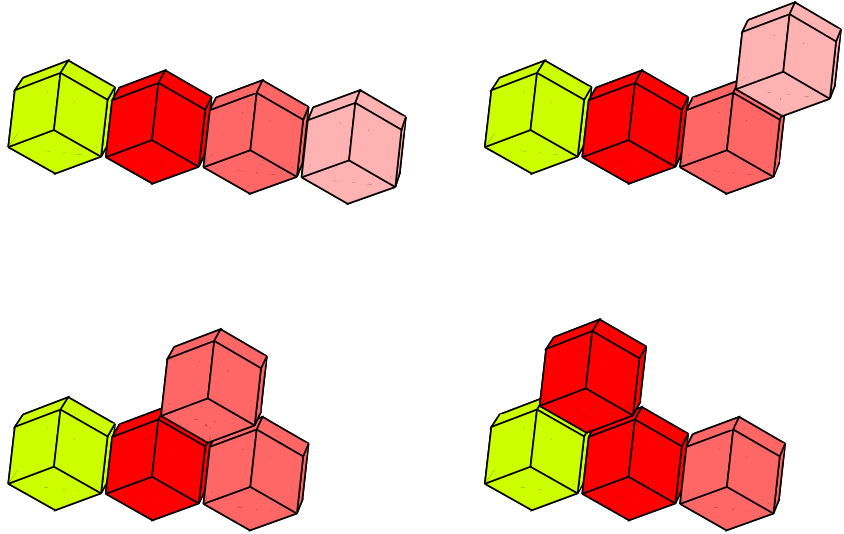,width=5in}
\end{center}
\caption{\label{search steps}Moving toward the seed. Starting from the initial configuration of Figure~\ref{search}, the module at the end of the chain moves toward the seed, reducing its scent value (shown by the colors) as it contacts other modules closer to the seed.}
\end{figure}

A simple example is shown in Figure~\ref{search}. A single SEED
module, with scent $S=0$, is connected to a chain of modules in SEARCH
mode with successively larger scent values. In this configuration,
only the SEARCH module at the end of the chain is free to move. Its
neighbor's smaller scent value (i.e., $S=2$) indicates that the module
at the end of the chain can find a position with even smaller scent
value by rolling over the faces of its neighbor. One possible sequence
of steps following from this initial configuration is shown in
Figure~\ref{search steps}.  For simplicity, this figure shows the
remaining modules remaining stationary.

In this example, the moving module goes directly toward the seed.
However, the gradient does not specify precisely which other face is
best. Instead, the free module moves randomly over the available faces
of its neighbor until it encounters the module with $S=1$. At this
point, the moving module updates its scent value to $S=2$ and then
continues its search for the seed by moving over the faces of the
$S=1$ module. Because there are only a few faces to explore before new
gradient information is found, this use of random search is likely to
complete after only a few moves.

\section{Results}

This section presents experiments on Proteo with some environmental
sensing and using our agent-based control algorithms within each
module. The experiments were restricted to no more than a few hundred
modules to keep simulation times reasonable (i.e., at most several
hours).

\subsection{A Chain}

A chain is often the preferred locomotion configuration of modular
robots for moving over steps, snaking into holes or squeezing through
narrow passages. We describe a simple local algorithm that, starting
from an arbitrary connected configuration, creates a single,
one-module-thick chain. Initially, all modules are in the SLEEP
mode. The other possible modes are: SEARCH, FINAL, and SEED. A module
is arbitrarily picked to serve as the initial seed to start the
chain. This initial seed can be determined centrally, or each module
in SLEEP mode can be given a small probability to become a seed at
each step. This module will pick a direction of growth and emit a
scent to start attracting other modules toward it. When a module moves
next to the seed module and is connected to it in the chosen
direction, the chain grows by one module and that new module then
becomes the new seed that will further grow the chain.

\begin{figure}
\begin{center}
\epsfig{file=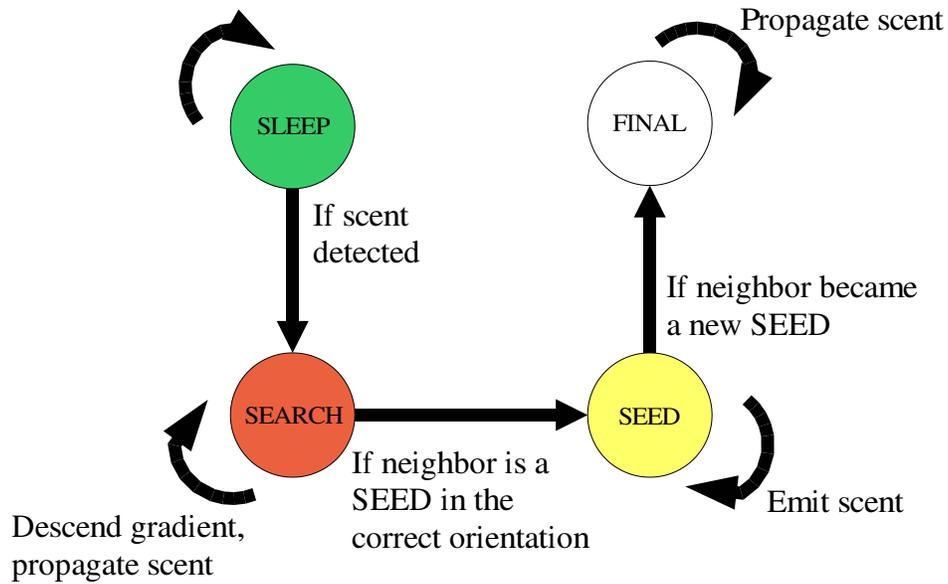,width=4in}
\end{center}
\caption{\label{chain flowchart}The module modes and conditions causing modules to change mode used to form a chain.}
\end{figure}

\begin{figure}
\begin{center}
\epsfbox{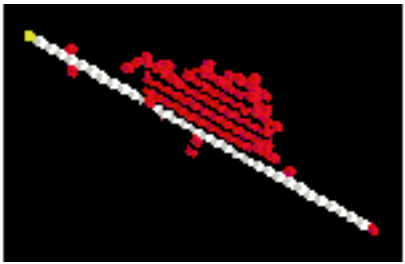}
\end{center}
\caption{\label{chain}
Chain Behavior. The Proteo system is shown here half-way to a complete
chain. The colors indicate the current module states: white modules have settled (FINAL mode) and red modules are still looking for a place in the structure (SEARCH mode). The yellow module is the current SEED.}
\end{figure}

The control rules for each module are:
\begin{itemize}
\item{If in SLEEP mode, if a scent is detected, go to SEARCH mode.}
\item{If in SEARCH mode, propagate scent and move along scent gradient.}
\item{If in SEED mode, emit scent, and if a module has appeared in the
direction of growth, set that module to SEED mode, and go to FINAL
mode.}
\item{If in FINAL mode, propagate scent.}
\end{itemize}
These rules amount to a simple finite-state machine for each module,
shown in Figure~\ref{chain flowchart}. An example configuration of a
group of modules following these rules is given in
Figure~\ref{chain}. As the algorithm continues, the modules form a
single chain.

\subsection{Recursive Branching for Locomotion and Manipulation}

A variety of applications for modular robots can benefit from a
branching structure with ``limbs'' at several levels. In such a
hierarchical structure, each level has varying degrees of precision,
range of motion and strength. For example, one level of branching
gives a structure that can be used as an artificial hand or as a
``spider'' for locomotion. Adding extra levels of branching results in
``fingers'' or ``toes'' with lower strength and range of motion but
higher level of precision. Human limbs exhibit only two levels of
branching, but a robot could grow as many levels as needed to achieve
increasingly complex manipulation and locomotion
tasks~\cite{moravec88}.

\begin{figure}
\begin{center}
\epsfbox{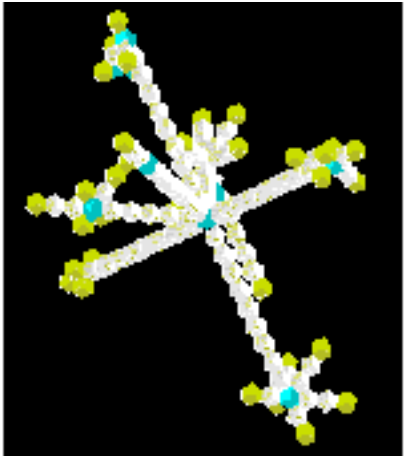}
\end{center}
\caption{\label{branching}
Branching Behavior. In this picture the second-level branches are not
completely grown because the available modules have been
depleted. Purple modules are nodes that spawn yellow seeds. The node
scent threshold was $T=12$ and maximum branch count $B=6$.}
\end{figure}

The algorithm is as follows. All modules are initially in the SLEEP
mode. Growth is initiated when a module randomly decides to switch to
the NODE mode. In this mode it attracts other modules (emits a scent)
and randomly chooses which ones will become seeds. When it has spawned
a certain number of seeds, the node becomes inactive. This experiment
uses two scents: a ``regular'' scent is used to grow the structure (as
in the Chain example), and a second ``node'' scent is used by modules
to determine how far the nearest node is. If this distance is too big,
a module can decide to become a node. The rules are:
\begin{itemize}

\item{If in SLEEP mode, if regular scent is detected, go to SEARCH mode.}
\item{If in SEARCH mode, propagate both scents and move along the regular
scent gradient.}
\item{If in SEED mode, propagate the node scent, and emit a regular scent;
if there is a neighboring module in the direction of growth of the
branch, set that module to be a seed, and go to FINAL mode.}
\item{If in FINAL mode, propagate both scents; if the node scent is weak
(i.e., has a value greater than some threshold $T$), go to NODE mode.}
\item{If in NODE mode, emit both scents, spawn seeds in random directions,
until a certain count is reached $B$, then go to INODE mode.}
\item{If in INODE mode, emit a node scent and propagate the regular scent.}

\end{itemize}
An example structure resulting from these rules is shown in
Figure~\ref{branching}.  The threshold $T$ determines the distance
between nodes, i.e., how far modules continue building one branch
before starting new branches. The count $B$ determines the branching
ratio throughout the structure.

\subsection{Dynamically Adapting to External Forces}

\begin{figure*}[t]
\begin{center}
\epsfbox{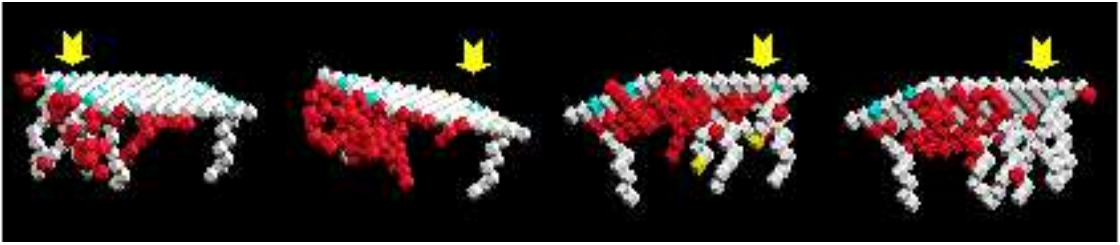}
\end{center}
\caption{\label{weight shift}
Adaptive Behavior, Sequence 1. Arrows indicate where the weight is
placed. Initially, the right side receives little stress, so it has
only one leg (leftmost picture). The weight is shifted to the right,
and the table adapts by disbanding most of the legs on the left, and
growing more legs on the right side (next three pictures). Red modules
are available for additional legs if needed. On the top of the table,
white modules are in FIXED mode while the blue ones are roots.  The
view in the last two pictures is from underneath the table to better
show the migration to the right side, and also shows some yellow SEED
modules. In this example, the the probability for a root to create or
disband legs are $\pmax=\pmin=0.05$. The weight thresholds were
$\Fmax=2$ to create new legs and $\Fmin=1$ to disband legs. In this
example, the board was a 30 by 30 square with roots distributed uniformly 
thoughout the board, separated by a distance of 3 modules in each direction.}
\end{figure*}

One important task for reconfigurable robots is to adjust themselves
in response to environmental forces. For instance, a collection of
such robots supporting a weight on a set of ``legs'' should be able to
change the location and density of the legs when the weight shifts. As
a simple example of this task, we examined the formation of legs to
support a flat structure with an additional imposed force whose
location could change over time. For simplicity, we neglect the
weights of the modules themselves, i.e., we assume their weights are
small compared to the weight of the additional objects they are
supporting on top of the flat structure or ``table''.

For this behavior, we form some of the modules into a ``board'' (the
top of the table) which supports the additional weight.  These
modules, which never move, are of two types. The first, always in
FIXED mode, only transmit scent.  The second type are ``root'' modules
which are uniformly spaced over the board and can communicate with
each other, e.g., through signals sent locally through the fixed
modules. We assume these signals propagate much faster than the
physical movements of the modules. The roots are grouped into regions
on the board (in the experiments reported here, we use two regions:
the two halves of the board).

We assume modules are equipped with a force sensor that allows them to
measure the weight they are supporting. Furthermore, we assume the
modules can either determine the direction gravity acts or this
direction toward the ground is prespecified when creating the roots in
the board. Roots within a region monitor the total weight supported in
that region by communicating their sensor readings.

The root modules on the board can be in one of three modes: ROOT,
IROOT or AROOT. Initially the root modules are in IROOT mode, which
monitor the weight they are supporting and may start growing a
leg. AROOT mode emits a scent to attract other modules to start
growing a leg, in much the same way as growing a single chain. Once
the leg starts growing, the root shifts to ROOT mode, where it remains
until the weight it supports drops below a specified threshold in
which case it probabilistically causes its leg to disband. Thus the
root modules grow or disband legs probabilistically according to how
much weight is experienced in their part of the table. We refer to
root modules in ROOT or AROOT mode as ``active'' roots.

The fixed and root modules are set at the beginning and don't change
throughout the experiment. All other modules are initially in SLEEP
mode.  The structure first grows legs, then disbands some of them and
creates new ones according to shifts in weight (controlled
interactively by the user of the simulator).

The rules for this behavior for the root modules are:
\begin{itemize}
\item{If in IROOT mode, if average weight per active root in its region is
above a certain threshold $\Fmax$, with a small probability $\pmax$:
go to AROOT mode.}
\item{If in AROOT mode, emit scent.}
\item{If in ROOT mode, if average weight per active root in its region is
below a certain threshold $\Fmin$, with a small probability $\pmin$:
set all FINAL and SEED neighbors to DISBAND mode, go to IROOT mode.}
\end{itemize}
Although similar adaptive behaviors can occur with a range of values
for the thresholds, one important issue is to prevent the disbanding
of a single leg supporting a minimal weight in a region. In our
experiments, this minimal value is one unit of weight, so we should
have $\Fmin \leq 1$.  The rules for the remaining modules are:
\begin{itemize}
\item{If in SLEEP mode, if a scent is detected, go to SEARCH mode.}
\item{If in SEARCH mode, propagate scent and move along its gradient; if
there is a neighboring SEED or AROOT module and location is towards
the ground with respect to this neighbor 
\begin{itemize} 
\item if this neighbor is a SEED, set it to FINAL mode, otherwise set it to
ROOT mode
\item go to SEED mode 
\end{itemize} }
\item{If in SEED mode, if touching the ground, go to FINAL mode.}
\item{If in FIXED or FINAL mode, transmit scent.}
\item{If in DISBAND mode, set all FINAL and SEED neighbors to DISBAND mode,
go to SEARCH mode.}
\end{itemize}

Figure~\ref{weight shift} illustrates how, once created, a supporting
structure can adapt in response to changes in external forces.  The
behavior dynamically adapts the location and number of its legs to
accommodate changes in the supported weight.  In this example, we
suppose weight $W_1$ is applied to the first half of the table and
$W_2$ applied to the second half, with an arbitrary choice of units
giving $W_1+W_2 = 10$.  In each half of the table, active roots
communicate their sensory information to determine how much of this
weight is supported by those in their half of the table. The result is
the average weight supported by each active root in each half of the
table, i.e., $w_1=W_1/N_1$ for the first half (where $N_1$ is the
number of active roots in the first half), and similarly for $w_2$ as
the average weight supported per active root on the second half.

The probabilistic growing and disbanding of legs avoids possible
oscillations in root behavior: if deterministic, when the weight
shifts, many roots could decide to grow legs, then disband them on the
next step, thus oscillating between the two without doing anything
useful. Such oscillations are common for systems with synchronous
updates~\cite{huberman93a,macready96}: randomization is a simple
technique to prevent spurious synchronization of agent activity.

After a root $r$ decides to grow a leg, some time is required for the
modules to move to that root and produce the leg. However, because a
root's decision to grow a leg is based on the weight in its region
averaged over the active roots, once root $r$ starts the process of
growing a leg by emitting scent, the average weight per active root
drops immediately and proportionately with the number of supporting
modules. This change, rapidly communicated among the roots, gives
feedback to other roots in the region and prevents other legs from
growing nearby unnecessarily. In practice, this implies that the
weight must be shifted slowly: the system will not be able to respond
quickly to sudden changes.

\begin{figure}[ht]
\begin{center}
\epsfbox{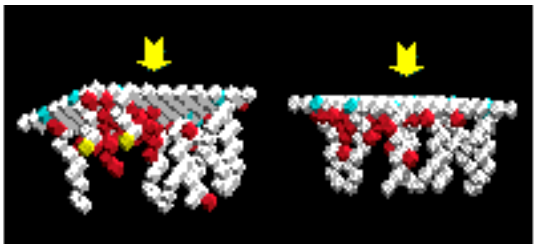}
\end{center}
\caption{Adaptive Behavior, Sequence 2. The weight is now moved to the center
of the table. The table adapts by shifting legs from the right to the
middle.}
\end{figure}

An indication of how fast the modules react to external conditions is
the number of steps after a weight shift until a new stable
configuration is achieved. In the examples shown here, this time is on
the order of 200 cycles. This includes the time needed to disband some
legs and construct some new ones.  Although the modules know about the
shifted weight immediately, some time is needed until some roots
decide to disband their legs, and some inactive roots decide to grow
legs. These decisions are probabilistic: each root decides to change
its state (go from active to inactive or vice versa) with a
probability we chose to be on the order of $1/R$ where $R$ is the
total number of roots in the structure. Thus it is unlikely that many
roots will decide to switch state at the same time step, thereby
avoiding unstable or oscillating system behavior.  We depend on the
rough assumption that decisions to disband or grow legs have immediate
effect on the strain felt by the modules in the system.  This
assumption is questionable for real-world applications, especially
with larger structures. However, it provides a good starting point. In
real-world applications the trade-off will be between fast response
with a risk of oscillating behavior and a slow, but stable response to
external conditions.

\subsection{\label{grasp}Grasping an Object}

The previous example showed how simple agent-based control allows the
robot to respond to changes in environmental stress. In practice, such
robots will be used to perform a globally specified task, leading to
the question of how high-level, imprecise specifications can be
combined with low-level behaviors.

As an example of such a task, we examined the behaviors required for
Proteo to reach out and grasp an object of unknown shape and size and
with only roughly specified location by ``growing'' around it. In this
behavior the only modules that do substantial work are the modules in
SEARCH mode. Initially all modules are in SLEEP mode, except for eight
modules that are set to SEED mode (to create eight branches to grow
toward the object). Two types of seeds are used: SEED modules, and
TOUCHSEED modules. Initially growth is caused by SEED modules, but
once the object is reached, seeds transfer to the TOUCHSEED state. The
two types of seeds allow modules in SEARCH mode to know whether to
grow in the direction of the object (to reach it) or to grow the
structure around the object (to grasp it). The system is initially
given the approximate direction the object is in, however, it has no
knowledge of the exact direction of the object, how far it is, or what
shape and size it has. The modules are assumed to have contact sensors
to detect when the object has been reached. The rules are simple:

\begin{figure}
\begin{center}
\epsfbox{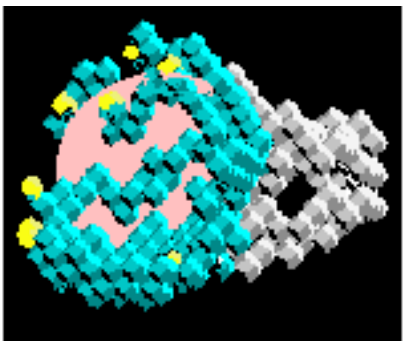}
\end{center}
\caption{Grasping Behavior. White modules (in FINAL mode) form eight 
``fingers'' that grow towards the object. When the object is reached,
the fingers start growing around it (with modules switching to TOUCH
mode, in blue). The modules have contact sensors to know if they are
touching the object. The yellow modules are the current TOUCHSEED's.}
\end{figure}

\begin{itemize}
\item{If in SLEEP mode, if scent is detected, go to SEARCH mode.}
\item{If in SEARCH mode, propagate scent, and move along the scent gradient;
if a seed has been found: if touching the object, then set seed to
TOUCH mode, and go to TOUCHSEED mode; otherwise, if not touching the
object, and seed is in SEED mode, find a spot next to the seed in the
direction of the object, set seed to FINAL mode, and go to SEED mode;
otherwise, make a random move.}
\item{If in TOUCHSEED or SEED mode, emit scent.}
\item{If in TOUCH or FINAL mode, propagate scent.}
\end{itemize}

This example incorporates a global goal, specified by the general
direction of the object, with local adjustments by the agents to the
actual object surface. It thus illustrates an important decomposition
of the control task into global specification in terms of general
parameters combined with low-level control based on the actual
environment of the modules. This decomposition considerably simplifies
the overall motion planning since the global specification need not
have access to all the details of the environment, knowledge of which
modules might have failed, etc.

These two examples could be combined: including stress sensors could
also allow the modules to adjust for the object weight if it is to be
picked up rather than just grasped.

\section{\label{stability}Local Minima and Stability}

Local control has the danger of getting trapped in locally
``comfortable'' configurations that are not globally suitable.  Our
agent approach provides good solutions to this problem both for
individual modules and the system as a whole.

A module searches for a seed by moving along a gradient of
least-descent. In Proteo, modules can only move by ``rolling'' over
the outside surface of the structure, while scents can propagate
through the inside of the structure. Usually this is not a problem:
all modules searching for seeds are constantly moving preventing any
fixed local extremum. However, a problem may arise when some modules
remain fixed. Modules searching for seeds may not be able to follow
the internal scent and get through to seeds on the other side because
fixed modules effectively block the way. There are two possible
solutions: one is to alter the design to allow modules to pass through
other modules (by squeezing through), and the other is to use only a
{\em surface scent}, which the modules can follow easily.

At a higher level, the whole system is performing a search for a
configuration that suits a given criteria. If the reconfiguration
rules are not set up carefully, however, the system may never converge
to a suitable state. For instance, if for each module we only specify
preferred immediate neighbor configurations, the most likely result is
that surface modules will settle quickly and comfortably, preventing
modules on the inside of the structure from moving. Thus, something
must guide the overall reconfiguration. We use the concepts of {\em
growth} and {\em seeds} to drive the reconfiguration of the whole
system in a focused way.

We found the structures grown are generally stable. During growth,
stability depends mostly on the ratio of scent propagation speed to
module speed. Since modules generally move slower than they can
transmit scent, the response to scents does not introduce oscillating
behavior~\cite{kephart89PhysRevA} or irregular growth, and the
intermediate configurations are well-balanced at all times.

\section{Conclusions}

We have presented an agent-based approach to self-reconfiguration for
modular robots that results in the creation of ``emergent'' structures
with the desired functionality. We use purely local, simple rules and
limited sensing. To ``grow'' stable structures, modules are guided by
local attractors (seeds) and global gradients (scents) toward good
configurations. Different structures result by varying the number and
combination of seeds and scents in the algorithm.

Because the method has a strong random component, the specific shape
of the resulting structures is non-deterministic. However, the
resulting structure has the desired functionality. This is
particularly well-suited to tasks with a strong element of uncertainty
in the precise nature of the task or in the environment, in contrast
to other reconfiguration algorithms.

We examined two tasks for which the reconfigurable nature of the
robots is well suited: adjusting structures to environmental stresses
and grasping an object whose detailed shape is not known. The global
behavior of the algorithms we developed for these tasks was determined
through simulations of the Proteo robot. The control primitives can
easily be scaled up to a larger number of modules, and down in the
size of individual modules, and are general enough to fit most modular
robot designs.

This approach could be used in conjunction with other
self-reconfiguration or control methods, as part of an overall
hierarchical control scheme to handle increasingly complex
tasks. Lately, an important research effort is applying genetic
algorithms to the control of autonomous systems~\cite{mitchell96} or
FPGA programming~\cite{thompson99}. The techniques described in this
paper are amenable to evolutionary approaches, and could benefit from
these results. Specifically, a population of agent programs combining
the control primitives in different ways could be tested against
variations in the desired task to evolve better behaviors according to
various performance metrics~\cite{bennett00}.  In the context of the
primitives presented here, genetic techniques could identify methods
to modulate the various parameters in our methods, e.g., the
probabilty to form a seed for new growth.

Finally, improvements in hardware design to augment current module
capabilities could greatly aid reconfiguration. Scaling down the size
of individual modules using micromachines
(MEMS)~\cite{bryzek94,berlin97} or molecular assemblies could benefit
actuation. For instance, protein motors can carry weights orders of
magnitude larger than their own, at speeds that are impressive given
their size~\cite{howard97,montemagno99}. This means a module could
carry several others around as it moves, something current Proteo
modules cannot do. In addition, designs that allow individual modules
to change their shape and size could also enhance overall
performance. These improvements, however, would require truly
interdisciplinary expertise in biology, materials, and electronics and
remain, for the most part, a formidable challenge. Nevertheless, as
such robotic systems are developed, multiagent control approaches are
likely to help them achieve robust behaviors.

\subsection*{Acknowledgments}
We thank Ying Zhang who wrote the original Proteo simulator used in
our experiments.


\end{document}